# Emotion : modèle d'appraisal-coping pour le problème des Cascades

**Karim Mahboub**[(1)], **Cyrille Bertelle**[(1)], **Véronique Jay**[(1)] &
**Evelyne Clément**[(2)]

[(1)]LITIS - Université du Havre (www.litislab.eu)
25, rue Philippe Lebon
BP 540 76058 Le Havre Cedex
{Karim.Mahboub, Cyrille.Bertelle, Veronique.Jay}
@litislab.eu

[(2)]Psy.Co – Université de Rouen
Rue Lavoisier
76821 Mont-Saint-Aignan
Evelyne.Clement@univ-rouen.fr

## 1. Modèles de l'émotion et appraisal-coping

L'étude des modèles émotionnels distingue deux principaux types d'approches : les modèles hiérarchiques et componentiels (Baudic & Duchamp, 2006). L'*appraisal-coping* (Smith & Lazarus, 1990) est un modèle componentiel dans lequel on distingue l'*appraisal* (évaluation cognitive) et le *coping* (réponse cognitive).

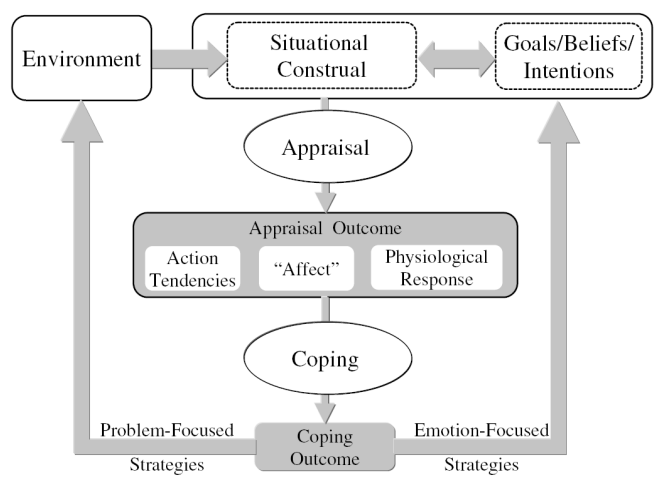

*Figure 1 : Système cognition-motivation-action,
adapté du modèle appraisal-coping de Smith et Lazarus (1990).*

L'*appraisal* consiste en l'évaluation cognitive de la situation par rapport à ses conséquences pour l'organisme (bien-être) et le *coping* aux stratégies d'ajustement qui orientent les réponses en fonction des conséquences évaluées.

Le système boucle sur une interaction réciproque de l'*appraisal* et du *coping*, traduisant l'interaction émotion-cognition-action.

## 2. Application au problème des Cascades

L'énoncé est le suivant : « Chaque case contient la somme des deux nombres situés au-dessus d'elle. Trouve les nombres qui manquent dans la grille ci-dessous. ».

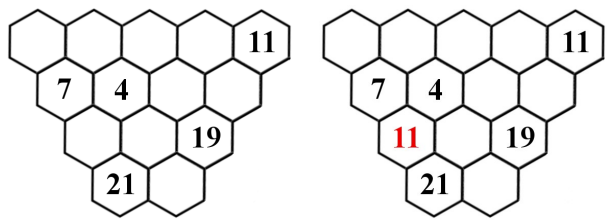

*Figure 2 : Grille initiale du problème des Cascades, accompagnée de la première étape de résolution.*

Proposée à des enfants de CM2, l'objectif de cette expérience est de pister les manifestations émotionnelles pendant le décours temporel de la tâche en enregistrant d'une part les expressions faciales et d'autre part les réponses électrodermales tout au long de l'expérience (Clément, 2007).

L'objectif du travail pluridisciplinaire réalisé jusqu'à présent est d'appréhender les liens entre émotion et cognition et de proposer un modèle de la cognition qui intègre la dimension émotionnelle. L'analyse des données issues des différentes expériences permet alors de mettre à l'épreuve les prédictions du modèle (Mahboub, 2006).

## 3. Modèle proposé

L'adaptation du modèle d'*appraisal-coping* au problème des Cascades nous permet de mieux identifier l'activité cognitive et émotionnelle des enfants. L'étape d'*appraisal* sert à l'évaluation et à la prédiction du plan sélectionné par l'enfant dans le but de résoudre le problème. En fonction de l'évaluation, il peut remplir un hexagone, en corriger un, ou bien si l'évaluation produit un mauvais résultat, il peut changer de plan (on observe que l'énoncé de l'exercice peut être interprété de différentes façons).

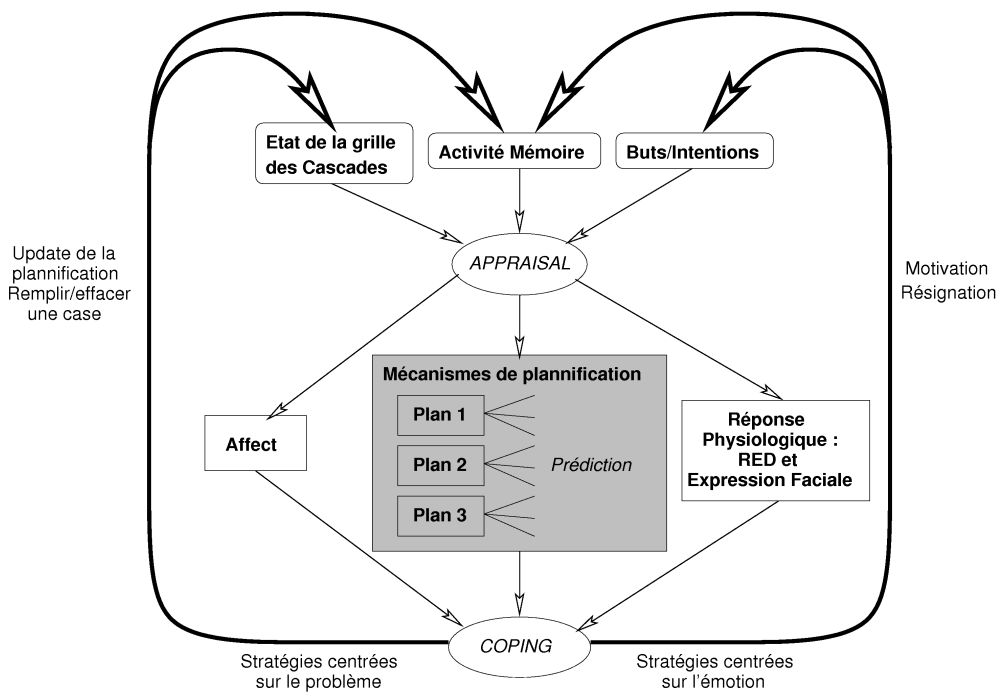

*Figure 3 : Modèle appraisal-coping appliqué au problème des Cascades*

La stratégie de *coping* est la décision effective que l'enfant va prendre en fonction de ses précédents choix. Elle est généralement accompagnée d'une réaction émotionnelle qui dépend des conséquences de l'évaluation. Par exemple, si l'enfant décide de confirmer son plan, son état émotionnel sera positif. A l'inverse, s'il change continuellement de stratégie, la situation peut se solder par un abandon de la tâche de résolution.

## 4. RÉFÉRENCES